# Adopting level set theory based algorithms to segment human ear

Bijeesh T. V[1] and Nimmi I. P[2]

Department of Computer Science and Engineering, Sree Narayana Guru College of Engineering and Technology, Payyanur, Kannur, Kerala, India

[1]`bijeeshtv@gmail.com`, [2]`nimmiip@gmail.com`

## ABSTRACT

*Human identification has always been a topic that interested researchers around the world. Biometric methods are found to be more effective and much easier for the users than the traditional identification methods like keys, smart cards and passwords. Unlike with the traditional methods, with biometric methods the data acquisition is most of the times passive, which means the users do not take active part in data acquisition. Data acquisition can be performed using cameras, scanners or sensors. Human physiological biometrics such as face, eye and ear are good candidates for uniquely identifying an individual. However, human ear scores over face and eye because of certain advantages it has over face. The most challenging phase in human identification based on ear biometric is the segmentation of the ear image from the captured image which may contain many unwanted details. In this work, PDE based image processing techniques are used to segment out the ear image. Level Set Theory based image processing is employed to obtain the contour of the ear image. A few Level set algorithms are compared for their efficiency in segmenting test ear images.*

## 1. INTRODUCTION

A Biometric is a measurable physical characteristic or a behavioral trait that can be used to authenticated a person. In this modern age of digital impersonation, authentication based on biometrics has become a more reliable measure against identity threats. The traditional authentication techniques such as keys and cards that the user has to carry with him always and the passwords and PINs the user has to remember always are fast becoming obsolete. Users find it difficult to carry the keys and cards or to remember the PINs. Moreover, forging keys, cards and PINs has now become the new trend among criminals. Thus an authentication mechanism based on biometric features is a necessity. Researchers across the world have widely studied physiological biometrics such as face and ear over the past few years. Both ear and face have been found to be an effective biometric to uniquely identify a human being. But biometric methods using ear holds an edge over face biometrics because of certain anatomical features of the ear.

Ear is one of our sensory organs and is therefore usually not hidden or covered by anything so that hearing is not affected. This makes data acquisition process much easier and efficient. In case where face is used as the biometric, data acquisition may be difficult due to spectacles, beard or make-up. Also, ear is a very stable anatomical feature in the human body. The features of human ear do not change drastically with time. Besides, ear is unaffected by emotions. On the other hand facial features change significantly with age. Facial features are also affected by various emotions that a human being can go through.

In this work, we propose to use level set theory based image processing techniques to obtain the contour of the ear image. Various algorithms that uses level set theory has been applied on the input ear image to segment the image and to obtain the ear contour. There are basically two types of segmentation methods that utilize level set theory. Edge based segmentation and Region based segmentation. We have experimented with both of these methods and a combination of Edge based and Region based methods. Features extracted from the contours obtained from these methods can be used to classify various ear images into predefined classes with the help of a classification algorithm such as Multiclass SVM. A combination of level set based segmentation techniques, feature extraction and a classification method can result in an extremely good automated Human Ear Recognition System.

## 2. RELATED WORKS

The first work on ear biometrics was done by Iannarelly in 1989 [1]. In his work, he proved the uniqueness of human ear by examining over 10000 ears. He established the uniqueness based on the distance between specific points of the ear. However the method was not automated as those specific points were extracted manually.

Research was focused on automating the extraction of key points from the ear image since then. Hurley *et al.* [2] introduced a method based on energy features of the image. They proposed to transform the image to force field and the energy lines, wells and channels are then extracted from the force field image. Another method proposed by Victor *et al.* [3], made use of PCA(Principal Component Analysis) to successfully identify human beings based on ear and face biometrics. Their method, however, was not fully automated, since the reference points had to be manually inserted into images. Another approach to ear image feature extraction was presented by Moreno *et al.* [4]. Their work was based on features extracted by compression networks. Several neural networks methods and classifiers combined are called Compression Networks. The success rate of 93% was achieved by the Compression Network ear identification method.

Yan and Bowyer [5] developed three approaches to 3D ear recognition problem, edge-based, ICP, and 3D-PCA. In their work, they used the energy map of the ear image to obtain the contour of the image, from which the features are extracted. They tested various approaches in multimodal biometric scenario. They designed fully automated ear recognition system and achieved satisfactory results of 97.6%. However, the ICP-based approach is still computationally expensive in comparison to other approaches considered.

Chen and Bhanu [6] proposed 3D ear recognition based on local shape descriptor as well as ICP algorithm. They used and ear detection approach that uses both color and range images to localize the ear region accurately by following a global-to-local registration procedure. Their results of ear detection, matching, and identification are close to 100% recognition rate.

M.Choras in his work 'Perspective Methods of human identification: ear biometrics, presented geometrical methods of feature extraction from ear images in order to perform human identification [7]. In order to perform human identification, he used geometrical parameters of the ear contour images. Contour extraction is done based on pixel illumination values and changes. Contours corresponding to earlobes are significantly diversified and contain enormous amount of information allowing ear identification. The feature extraction methods used in this work are concentric circles based method (CCM), contour tracing method (CTM), angle based contour representation method (ABM) and methods based on geometric parameters (GPM). Their method when tested under controlled environment with cooperative users, yielded results close to 100%.

## 3. PROPOSED METHOD AND ITS SCOPE

In this work, we propose a new method to segment the ear image and obtain the contour of the ear image. Various level set based segmentation algorithms are used to obtain the contour of the image. A level set is an implicit representation of a curve. In level set based segmentation, we first arbitrarily fix our level set on the image and then evolve the level set according to some force. Usually the force that controls the level set evolution is the curvature of the level set function, which is a geometric property of the function. We use a special function called edge stopping function which stops the level set evolution at the image boundaries. Edge stopping function is usually a function of the image gradient. Two types of level set based segmentation algorithms are edge based methods and region based methods. In our work, we have used both edge based and region based methods and sometimes a combination of both is used.

PDE (partial differential equation) based image processing is the latest and growing trend in image processing area. Segmentation based on level set theory is very simple and efficient. We can obtain the contour of the ear image with very minimal preprocessing steps. The obtained contour contains enormous amount of information which can help in ear identification. Once the contour of the image is obtained, we can extract important features from it. This level set based algorithm combined with a good feature extraction method and a classification algorithms can lead to a high accuracy automated human identification system.

### 3.1 Level set theory based segmentation

We begin by choosing a level set function, which generally is an implicit representation of a curve. Initially we fix the contour C0 arbitrarily (called the initial contour) and evolve it according to some force. An image property is used as the force that drives the level set evolution, which usually is the image gradient.

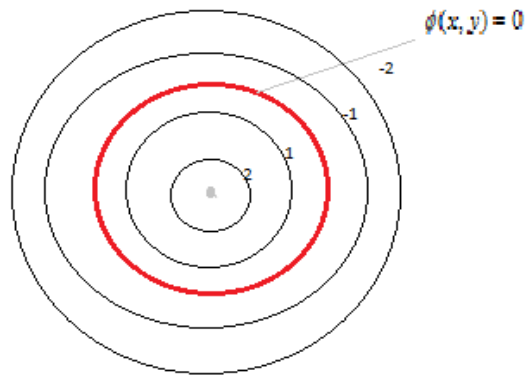

Figure 1: A level set function

Let the level set function be; $\phi(x, y, t)$ where t is an artificial time variable we introduce. The evolution of the level set function is governed by the equation,

$$\frac{\partial \phi}{\partial t} = F |\nabla \phi| \qquad (1)$$

The function $\phi(x,y,t)=0$ corresponds to the initial contour C0. The driving force F acts in a direction normal to the contour. Usually in a curve evolution model, the driving force F is the curvature of the level set curve, which is mathematically defined as follows.

$$F = \frac{u_y^2 u_{xx} - 2u_x u_y u_{xy} + u_x^2 u_{yy}}{\left(u_x^2 + u_y^2\right)^{3/2}} \qquad (2)$$

And $\nabla \phi$ is the gradient of the level set function $\phi(x,y,t)$.

This means that the level set evolution is controlled by two forces; the curvature of the function, which is a geometric quantity and the gradient of the level set function. The evolution continues until it can no longer evolve.

## 3.2 Image segmentation using level set methods

To perform image segmentation using level set evolution method, we add a new term called edge stopping function the level set evolution PDE. So the new PDE becomes,

$$\frac{\partial u}{\partial t} = g(x,y) K |\nabla u| \qquad (3).$$

Here the function $g(x,y) = 1/\left(1 + |\nabla f|^2\right)$, where $f$ is the image, is called the edge stopping function. The value of this function becomes nearly equal to zero at the object boundaries because of the fact that the image gradient is relatively higher at the boundaries. Therefore, $\frac{\partial u}{\partial t}$ becomes zero at the boundaries and thus the level set evolution stops.

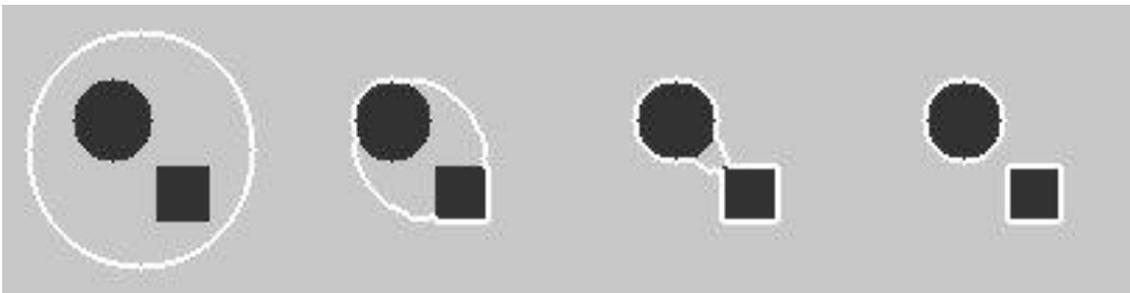

Figure 2: Image segmentation using level set evolution

Based on active contour techniques, we have two models for boundary detection, edge based and region based models: edge based models and region based models.

### 3.2.1 Edge based model

Edge-based methods usually use a measure of the changes across an edge, such as the gradient or other partial derivatives. Such methods utilizes image gradient to construct an edge stopping function (ESF) to stop the contour evolution on the object boundaries.

An edge stopping function may be represented by:

$$g(x,y) = \frac{1}{1+\left|\nabla(G_\sigma * f(x,y))\right|^2} \quad (4)$$

Here $G_\sigma$ is a Gaussian filter used to diffuse or smear the edges. The function $g(x,y)$ slows down the evolution of the level set curve at the edges. It has value close to zero near the edges, because the gradient of the image will have a high value at the image boundaries as there is a sharp difference in the pixel values at the boundaries.

But edge based method is not considered the best as the edges may not be sharp due to fading of ink or degradation which can prevent the gradient value at the edges from being a high value. Another disadvantage of these models is that they are very sensitive to the location of the initial contour. Edge base models are said to have local segmentation property as they can segment the objects only if the initial contour is placed surrounding the object.

### 3.2.2 Region based model:

This model does not work on the discontinuity in the image, rather it partitions the image into "objects" and "background" based on pixels intensity similarity. Region based models utilize the statistical information inside and outside the contour to control the evolution, which are less sensitive to noise and have better performance for images with weak edges or without edges.
In a region based segmentation model, we try to minimize the following energy:

$$\min_{C,c1,c2} E(C,c1,c2) = \int_{inside(c)} (f-c_1)^2 \, dxdy + \int_{outside(c)} (f-c_2)^2 \, dxdy + \lambda Length(C) \quad (5)$$

Here c1 and c2 are the average pixels values inside and outside the contour respectively.
The value of this integral is minimized when the contour C is on the edges of the objects to be segmented.
This method is not very sensitive to the location of the initial contour and can detect interior and exterior boundaries at the same time. Therefore region based models are said to posses' global segmentation property.

### 3.3 Algorithms used for segmentation

### 3.3.1 Distance regularized level set function (DRLE)

In conventional level set formulations, level set functions eventually develops irregularities during the evolution, which may result in unstable evolution. This problem is usually countered using a numerical remedy called re-initialization. Re-initialization periodically replaces the faulty or degraded level set function with a signed distance function which will act as the new level set function. Re-initialization raises challenges as to when and how it should be done and also results in numerical inaccuracy.

DRLS provides a new variation formulation which intrinsically ensures the regularity of the level set function. The level set evolution is derived as the gradient flow that minimizes the energy functional with a distance regularization term and an external energy that drives the motion of the zero level set toward desired locations. The distance regularization term is defined with a potential function such that the derived level set evolution has a unique forward-and-backward (FAB) diffusion effect, which is able to maintain a desired shape of the level set function, particularly a signed distance profile near the zero level set. The distance regularization avoids the need for re-initialization and thereby eliminating the inherent numerical inaccuracy. It allows the use of a more general level set function and while implementing, relatively high time step can be used which will reduce the number of iterations thereby reducing the computational cost.

### 3.3.1.1 Energy formulation with distance regularization

Let $\phi : \Omega \to \Re$ be a level set function defined on a domain $\Omega$. We define the energy functional $E(\phi)$ by,

$$E(\phi) = \mu R_p(\phi) + E_{ext}(\phi) \tag{6}$$

Where $R_p(\phi)$ is the level set regularization term defines as,

$$R_p(\phi) = \int_\Omega p(|\nabla \phi|) d\Omega \tag{7}$$

Where p is the potential (energy density) function. And $E_{ext}(\phi)$ is the external energy function which is designed such that it achieves a minimum when the zero level set of the LSF is located at desired position. The purpose of imposing the level set regularization term is to smooth the level set function $\phi$, at the same time it should also maintain the signed distance property $\nabla \phi = 1$ at least in the vicinity of the zero level set, in order to ensure accurate computation for curve evolution. This can be achieved by using a potential function $p(s)$ with a minimum point s=1, such that the regularization term $R_p(\phi)$ is minimized when $\nabla \phi = 1$. Therefore, the potential function can be defines as

$$p(|\nabla \phi|) = \frac{1}{2} \int_\Omega (|\nabla \phi| - 1)^2 d\Omega \tag{8},$$

which characterizes the deviation of $\phi$ from a signed distance function.

### 3.3.2 Region-scalable fitting energy minimization

Region-based models aim to identify each region of interest by using a certain region descriptor to guide the motion of the active contour. Most of the region based models depend on the intensity homogeneity of the regions to be segmented. But intensity in homogeneity often occurs in real world images and this makes segmentation a difficult task. In order to surmount this difficulty a new region-based active contour model was developed which relies on the intensity information in local regions. Region-Scalable Fitting (RSF) energy is defined in terms of a contour and two fitting functions that locally approximate the image intensities on the two sides of the contour. The optimal fitting functions are shown to be the averages of local intensities on the two sides of the contour. This is then incorporated in to variational level set formulation with a level set regularization term, from which a curve evolution PDE is derived for energy minimization.

The local intensity fitting energy is for a given point $x \in \Omega$, is defined as

$$E_x^{Fit}(C, f_1(x), f_2(x)) = \sum_{i=1}^{2} \lambda_i \int_{\Omega_i} K(x-y)|I(y) - f_i(x)|^2 dy \tag{9};$$

where C is the contour, $\lambda_1$ and $\lambda_2$ are positive constants, $f_1(x)$ and $f_2(x)$ are approximate image intensities inside and outside C. I is the image and K is a kernel function defines as, $K: \Re^n \to [0, +\infty)$ and has the following properties.

1. $K(-u) = K(u);$
2. $K(u) \geq K(v)$, if $|u| < |v|$ and $\lim_{|u| \to \infty} K(u) = 0;$
3. $\int K(x) dx = 1$.

The kernel function and its localization property (the second property) play a key role in this method. The size of the region can be controlled by the kernel function K and thus the local intensity fitting energy is otherwise called region-scalable fitting (RSF) energy.

$E_x^{Fit}$ is a weighted mean square error of the approximation of the image intensities inside and outside the contour C by fitting values $f_1(x)$ and $f_2(x)$ respectively, with $K(x-y)$ as the weight assigned to each intensity.

Because of the inclusion of a regularization term the regularity of the level set function is maintained intrinsically and the computational cost is reduced.

### 3.3.3 Localizing Region-based active contours

Region based energy is reformulated in a local way in this method. For images with heterogeneous intensity backgrounds, contour evolution using local image statistics works better than the methods using global segmentation. This method can be used with any global region based energy and bring about the advantages of localization as well. In this model, the foreground and the background are described as smaller local regions, removing the assumption that foreground and background can be represented using global statistics. These local regions lead to construction of local energies at each point along the curve. To compute these local energies, local regions are divided into local interiors and local exteriors by the evolving curve. The energy optimization is then done by fitting a model into each local region.

Let $I$ be an image defined on the domain $\Omega$ and C be the closed contour which is the zero level set of signed distance function $\phi$. The interior of C is defined by the following smoothed Heaviside function.

$$H\phi(x) = \begin{cases} 1, & \phi(x) < -\varepsilon \\ 0, & \phi(x) > \varepsilon \\ \frac{1}{2}\left\{1 + \frac{\phi}{\varepsilon} + \frac{1}{\pi}\sin(\pi\phi(x))\right\}, & \text{otherwise} \end{cases} \quad (10)$$

Similarly the exterior of C is defined as $(1 - H\phi(x))$.

To specify the area just around the curve a derivative of $H\phi(x)$ is used, which is a smoothed Dirac delta function.

71

$$\delta\phi(x) = \begin{cases} 1, & \phi(x) = 0 \\ 0, & |\phi(x)| < \varepsilon \\ \frac{1}{2\varepsilon}\left\{1 + \cos\left(\frac{\pi\phi(x)}{\varepsilon}\right)\right\}, & \text{otherwise} \end{cases} \quad (11)$$

We use another function $B(x, y)$ to mask the local regions. This function will be 1 when the point is within a ball of radius $r$ centered at $x$, and 0 otherwise.

$$B(x, y) = \begin{cases} 1, & \|x - y\| < r \\ 0, & \text{otherwise} \end{cases} \quad (12)$$

Here $y$ is another spatial variable in $\Omega$, which is independent of $x$.

The energy is given as follows,

$$E(\phi) = \int_{\Omega_x} \delta\phi(x) \int_{\Omega_y} B(x, y) F(I(y), \phi(y)) dy dx \quad (13)$$

where the function $F$ is a generic internal energy measure used to represent local adherence to a given model at each point along the contour. For computing $E$, we consider only the contributions from the points near the contour. This model ignores the non-homogeneity that may occur far away and thus results in a better segmentation of objects with non-homogeneous intensity backgrounds.

## 4. RESULTS AND DISCUSSIONS

The segmentation algorithms described in the previous section has been applied on various ear images obtained using a digital camera. We have restricted our experiments only to images taken in similar lighting conditions. The images were taken in a well illuminated room. Also we have excluded ears with occluding hair from our experiments.

The algorithms DRLE, RSF energy minimization and Localizing region-based active contour are separately tested on each ear image. All the processing is done on gray scale version of the image. So each algorithm first converts the colour image into a gray scale image. Also, the image is rescaled to reduce the size in order to reduce the time taken for the completion of the algorithms.

### 4.1 Segmentation using DRLSE

Distance regularized level set evolution method provide a satisfactory segmentation of the ear. The ear lobe was segmented properly but the subtle folds in the ear were not segmented. Since the earlobe geometry can provide good features for classification, this method can very well be used for creating an ear recognition system.

The desired segmentation result was obtained after various attempts using different initial contour shapes and locations. The initial contour used in our experiments is shown in the following figure.

International Journal on Cybernetics & Informatics ( IJCI) Vol.2, No.4, August 2013

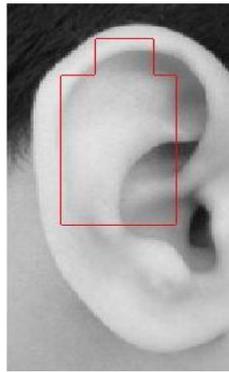

Figure 3: Initial contour

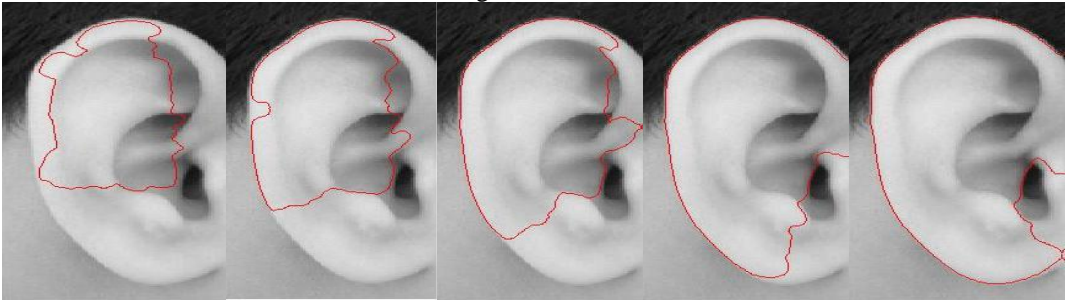

Figure 4: contour evolution

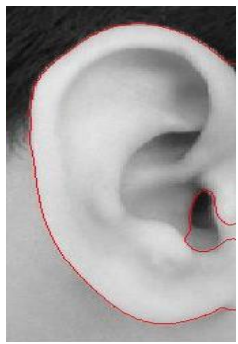

Figure 5: Final contour

As mentioned before, the earlobe is correctly segmented from the rest of the region. The contour around the earlobe can be extracted and various geometrical features can be extracted from the contour. Also, there is a small contour formed near the ear pit region. This can either be extracted and used for feature extraction or can be discarded if found to carry no meaningful information for ear characteristics. The algorithm took 860 iterations to evolve the level set to its final location and the time taken for contour generation was found to be 65.5582 seconds.

Figure 6 shows ten test ear images and the corresponding contours formed around them.

International Journal on Cybernetics & Informatics ( IJCI) Vol.2, No.4, August 2013

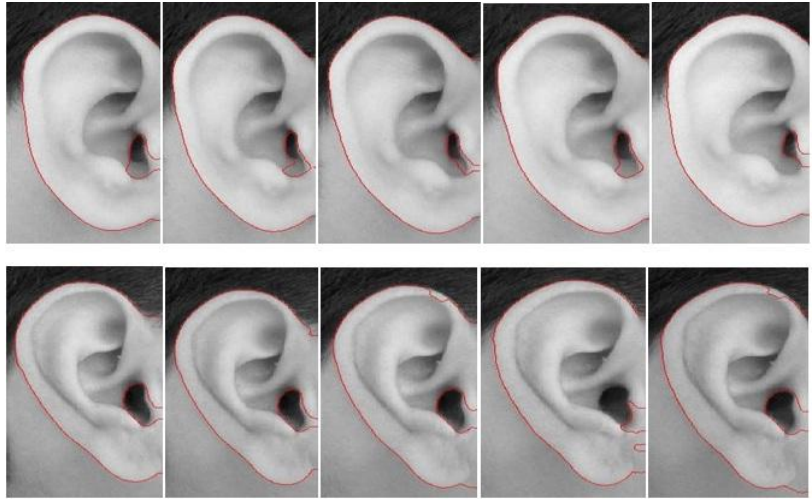

Figure 6: Contour formed around different input ear images

## 4.2 Segmentation using RSF energy minimization algorithm

Region scalable fitting energy minimization was faster than DRLSE algorithm in generating the contour. The segmentation algorithm was able to detect the earlobe and the many of the folds of the ear that can give important features for ear identification.

The initial contour and the contour evolution on the image are illustrated in the following figures.

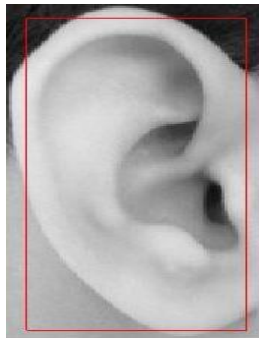

Figure 7: Initial contour

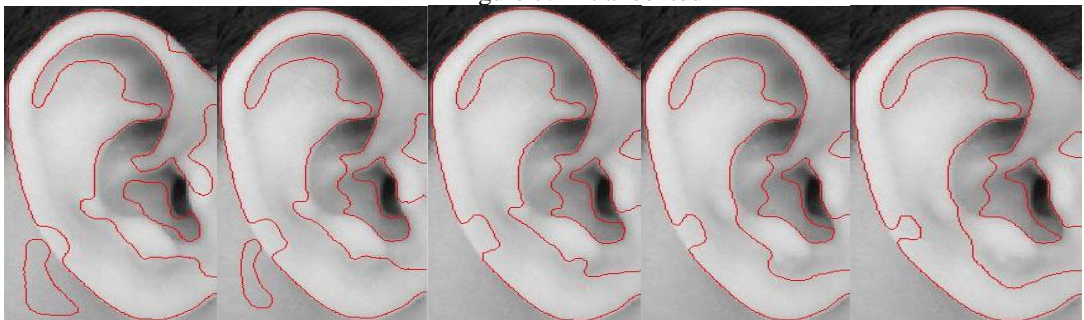

Figure 8: Contour evolution

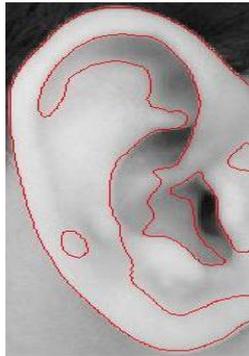

Figure 9: Final contour

This algorithm also was able to segment the earlobe correctly from the rest of the image. It also has detected various folds in the ear partially. The algorithm exhibited satisfactory level of consistency in segmenting with various ear images of different subjects. The main problem with this algorithm is that, it resulted in some extra unwanted contours. But these contours are small in size and can easily be removed or discarded by applying a thresholding on contour length. The algorithm was able to get the final contour in 300 iterations. The time taken for algorithm to segment the image and return the contour was learned to be 26.2234 seconds.

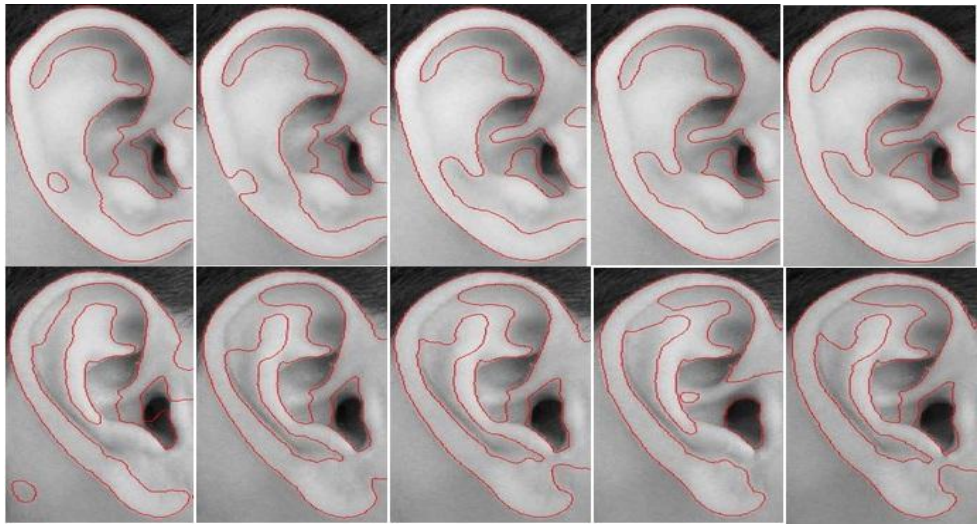

Figure 10: Input ear images and contour around them.

## 4.3 Segmentation using Localizing Region-based active contours

Localizing Region-based active contours method also was successful in segmenting the earlobe from the rest of the image. The folds in the ear were not detected by the algorithm. The initial contour shape and location were fixed after testing with various positions and shapes. The initial contour, contour evolution and the final contour are shown in the following figures.

75

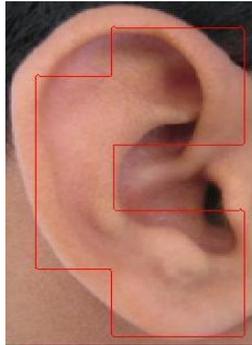

Figure 11: Initial contour

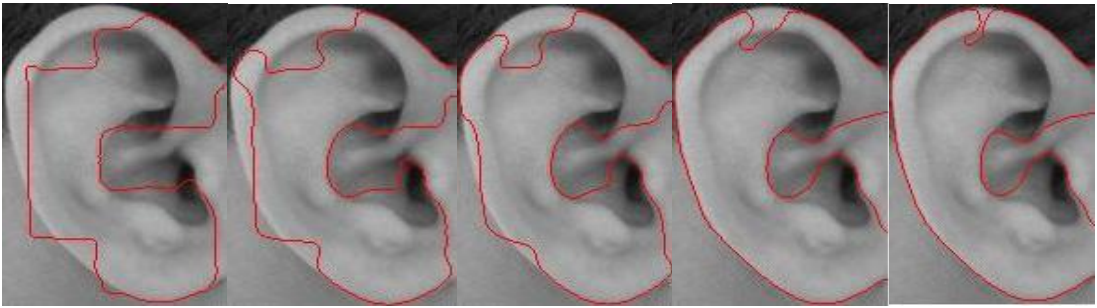

Figure 12: Contour evolution

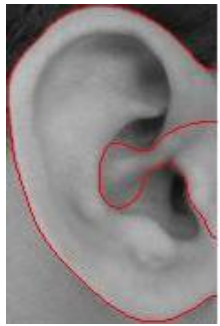

Figure13: Final contour

As can be observed from the figure, this method too was successful in segmenting the earlobe. There is an extra contour formed near the ear pit region. It partially detects the folds near the ear pit, but fails to segment it properly. Further research has to be done with the initial contour placement to obtain better segmentation results. This method took more time to process the image and generate the contour. It ran up to 940 iterations before it could evolve the contour to its final position. And it took as many as 103.4523 seconds to complete the algorithm. The algorithm behaved inconsistently with ear images of different test subjects. Further research has to be done to bring in more consistency. A set of input images and their corresponding contour images can be seen in the following figure.

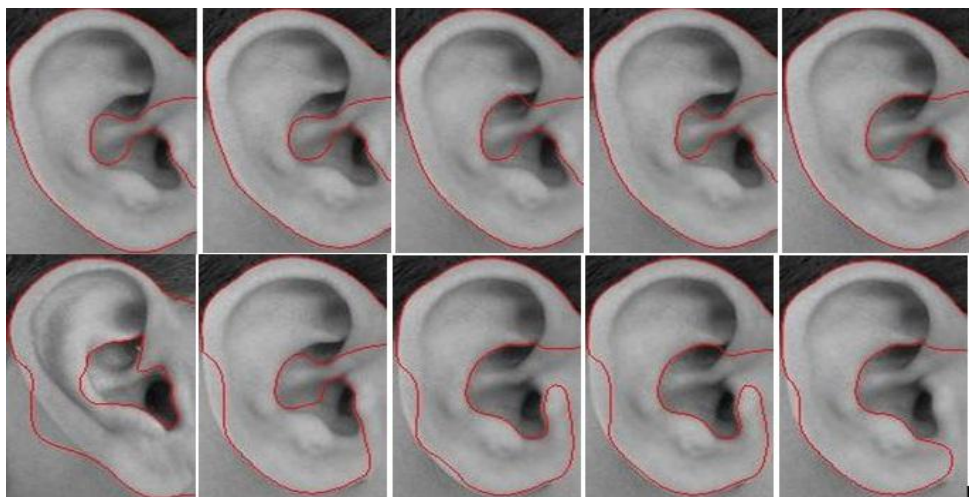
Figure 14: Contour formed around different ear images

All the above discussed segmentation algorithms can be used together to get all the relevant contours that can provide us with information which can be used in identifying an ear. After applying these algorithms and obtaining all the contours, we must have a means to discard the unwanted contours. This can be done mainly by discarding contours with length less than a certain threshold value.

### 4.4 Drawbacks of the proposed method

Main drawback of using the proposed methods for segmenting human ear is that the images should be taken in a steady environment where the lighting conditions remain the same throughout the data acquisition stage. Any region which is too bright or too dark will be segmented regardless of their importance in identifying an ear image. Also, earrings and occluding hair pose another major challenge. If occluding hair is present, the contour evolution breaks at that point and eventually results in undesired contours as output. Furthermore, since the folds that are present in the ear are not very fine edges and do not vary much in pixel intensities, it is quite hard to detect them using these methods.

Another problem is fixing the number of iterations for the algorithms. Although we use an edge stopping function to stop the contour evolution at the edges, since many of the edges in the ear image do not have a high variation in pixel intensities, the level set at times tends to move further away from the edges. Because of this problem the number of iterations up to which the algorithm should run has to be fixed properly. Also fixing the position and the shape of initial contour can be another time consuming job.

## 5. CONCLUSION

Human identification based on biometric features has now become very popular. Among other biometrics features, human ear was found to be a very good candidate for uniquely identifying a human being. The first step in developing an automated ear recognition system based on image processing techniques is to separate out the ear portion from the rest of the image. We, in this paper, have presented a method to segment out the ear image from the rest of the image and our method gives the contour image of the input ear image as the output. Features that can uniquely identify human ear can be extracted from this obtained contour image. We have used level set based segmentation techniques to obtain the contour of the ear image. Algorithms used in this work are combination of edge based and region based models, and thus imbibes the advantages of both methods. The experiments were conducted using mainly three algorithms, which are Distance regularized level set evolution, region scalable fitting energy minimization and localizing region-based active contours. Level set based methods hold a clear advantage over other segmentation methods as it is very easy to implement.

## 6. FUTURE WORK

So far in this work we have used only a single initial contour that evolved itself to form the final contours. The next improvement can be obtained by placing multiple initial contours. This might help in detecting all the relevant edges. We have proposed a method to segment the human ear images. Features can be extracted from the obtained contours and along with a good classification method; a fully automated human ear recognition system can be developed.

**Authors**

**Bijeesh T.V** received the B.Tech degree in Computer Science and Engineering from Kannur University, Kerala, India, in 2006, the M.Tech degree in Computational Engineering and Networking from Amrita Vishwavidyapeetham, Coimbatore, Tamilnadu, India, in 2012. He was a Project Engineer at Wipro Technologies for 3 years. He is currently working as an Assistant Professor in CSE department, Sree Narayana Guru College of Engineering and Technology, Kerala, India. 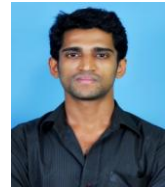

**Nimmi I. P** received the B.Tech degree in Computer Science and Engineering from Kannur University, Kerala, India, in 2009, the M.Tech degree in Cyber Security from Amrita Vishwavidyapeetham, Coimbatore, Tamilnadu, India, in 2012.She is currently working as an Assistant Professor in CSE department, Sree Narayana Guru College of Engineering and Technology, Kerala, India. 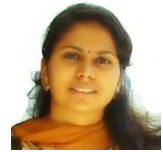